\documentclass[runningheads]{llncs}
\usepackage[T1]{fontenc}
\usepackage{amsmath}
\usepackage{graphicx,verbatim}
\usepackage{textcomp}
\usepackage{diagbox}
\usepackage{multirow}

\usepackage{tabularx,booktabs}
\newcolumntype{P}[1]{>{\centering\arraybackslash}p{#1}}
\usepackage[
colorlinks=true,
urlcolor=blue,
linkcolor=black,
citecolor=black
]{hyperref}
\newcolumntype{Y}{>{\centering\arraybackslash}X}

\newcommand{\equalcontrib}{\textsuperscript{$*$}} 
\newcommand{\lastauth}{\textsuperscript{$\star$}}

\usepackage{color}

\begin{document}
\footnotetext{\equalcontrib~Equal contribution}
\footnotetext{\lastauth~Co-corresponding: \texttt{\{carsten.marr,ario.sadafi\}@helmholtz-munich.de}}
\title{Attention Pooling Enhances NCA-based Classification of Microscopy Images}
\titlerunning{Attention Pooling Enhances NCA-based Classification of Microscopy Images}

\author{Chen Yang\inst{1,2}\equalcontrib  \and
Michael Deutges\inst{1}\equalcontrib  \and
Jingsong Liu \inst{3,4} \and
Han Li \inst{2,3} \and
Nassir Navab \inst{2} \and \\
Carsten Marr \inst{1}\lastauth \and Ario Sadafi \inst{1,2}\lastauth}

\authorrunning{C. Yang and M. Deutges et al.}


\institute{Institute of AI for Health, Computational Health Center, Helmholtz Munich, Munich, Germany \and
Computer Aided Medical Procedures, Technical University of Munich, Munich, Germany \and
Institute of Pathology, Technical University Munich, Munich, Germany \and
Munich Center for Machine Learning, Munich, Germany }

\maketitle              
\begin{abstract}
Neural Cellular Automata (NCA) offer a robust and interpretable approach to image classification, making them a promising choice for microscopy image analysis. However, a performance gap remains between NCA and larger, more complex architectures. We address this challenge by integrating attention pooling with NCA to enhance feature extraction and improve classification accuracy. The attention pooling mechanism refines the focus on the most informative regions, leading to more accurate predictions. We evaluate our method on eight diverse microscopy image datasets and demonstrate that our approach significantly outperforms existing NCA methods while remaining parameter-efficient and explainable.
Furthermore, we compare our method with traditional lightweight convolutional neural network and vision transformer architectures, showing improved performance while maintaining a significantly lower parameter count. Our results highlight the potential of NCA-based models an alternative for explainable image classification.


\end{abstract}
\section{Introduction}
The accurate classification of cells in microscopy images is essential for medical diagnostics, aiding in the diagnosis and monitoring of various diseases. Traditionally, this task relies on medical experts, but manual analysis is costly, time-consuming, and susceptible to human error. Deep learning has significantly advanced this field, promising to provide an efficient, consistent, scalable, and accessible solution. \cite{hehr2023explainable,KARRI2022797,liu2025hasd,matek2019human,rajaraman2018understanding,sadafi2023redtell,sadafi2019multiclass,diagnostics13071299}


However, state-of-the-art architectures such as Convolutional Neural Networks (CNNs) and Vision Transformers (ViTs) often have large parameter counts and require extensive training data to achieve high performance. In low-data medical imaging settings, these models are prone to overfitting and may struggle to generalize across domains. While lightweight variants have been proposed \cite{howard2019searching,mehta2021mobilevit}, they often trade off accuracy for reduced complexity, leaving a gap between model capacity and generalization.

A promising alternative to overcome these limitations are Neural Cellular Automata (NCA) \cite{mordvintsev2020growing}. NCA iteratively update feature maps through local interactions, enabling information propagation over multiple steps. This iterative process allows for effective feature extraction while maintaining a minimalistic design.

A key limitation in previous NCA-based classification models is the transition from high-dimensional feature maps to compact feature embeddings. Deutges et al. \cite{deutges2024neural} proposed an NCA model for classification, where a simple maximum pooling operation is used to reduce dimensionality. However, this approach can result in the loss of valuable information. Tesfaldet et al. \cite{tesfaldet2022attention} explored the integration of self-attention mechanisms within an NCA framework for denoising, showing the potential of spatial attention in NCA-based models. 

Here, we introduce attentionNCA (aNCA), a novel NCA-based image classification model that integrates attention pooling to enhance feature aggregation. In contrast to the simple maximum pooling proposed by Deutges et al.~\cite{deutges2024neural} our approach enhances NCA with a spatial attention pooling mechanism. This learnable, weighted aggregation method focuses on the most informative regions of the feature maps, improving representation quality and classification performance while maintaining a lightweight architecture with only 89k parameters. In contrast to the work of Tesfaldet et al.~\cite{tesfaldet2022attention}, our method differs significantly in the way the attention mechanism is integrated and specifically focuses on enhancing feature aggregation for classification tasks.

We evaluate aNCA across eight diverse microscopy datasets, including white blood cells, Pap smears, urine sediments, malaria infected cells, and pathology tissues. Our model consistently outperforms existing NCA-based classification models and conventional lightweight architectures. To foster reproducible research, we publish our source code at \url{https://github.com/marrlab/aNCA}.

\section{Methods}
We propose a classification network that utilizes an NCA backbone for feature extraction. To aggregate the extracted features, we introduce a learnable spatial attention mechanism that emphasizes the most informative regions. The highest 10\% of pixel values across each feature channel are then averaged to produce a compact, low-dimensional feature embedding. This embedding is passed through a fully connected classifier to generate the final prediction.

\subsection{NCA backbone}
NCA operate through iterative updates, where each cell's state is modified based on its local neighborhood \cite{deutges2024neural}.
Here, a cell refers to an individual pixel across all feature channels. 

The update rule for a given cell $c$ is defined by a transition function that takes the cell’s local $3\times 3$ neighborhood $N_c$ as input and computes an update. 
Firstly, information from the local neighborhood is aggregated by the perception function $f_p$, which applies two convolutional filters ($\kappa_1$ and $\kappa_2$) to the neighborhood. The resulting features are concatenated with the cell's current state $c^t$ to form the perception vector ($f_p(N_c)$). The update function $f_u$ then processes this information through a two-layer fully connected network with Rectified Linear Unit (ReLU) activations, parameterized by weights 
$\Tilde{W}_1$, $\Tilde{W}_2$ and biases $\Tilde{b}_1$, $\Tilde{b}_2$:
\begin{align} 
&f_p: \mathbf{R}^{3\times 3 \times n} \rightarrow \mathbf{R}^{3n}, && N_c \mapsto \left(c, N_c * \kappa_1, N_c * \kappa_2 \right)^T,\\ 
&f_u: \mathbf{R}^{3n} \rightarrow \mathbf{R}^{n}, && f_p(N_c) \mapsto \Tilde{W}_2 \max(\Tilde{W}_1 f_p(N_c) + \Tilde{b}_1, 0) + \Tilde{b}_2. 
\end{align}
Finally, the output of the update function is added to the current state:
\begin{equation} 
    c^{t+1} = c^t + \delta f_u(f_p(N_c)). 
\end{equation}
On average, only half of the cells are updated per iteration, due to the stochastic binary variable $\delta$, 
which acts as a regularization mechanism, enhancing robustness.

At a given time step $t$, the full set of cell states is represented as $S_t \in \mathbf{R}^{H\times W\times n}$ for an image with height $H$, width $W$, and $n$ feature channels. A single NCA update across the entire image can therefore be expressed as
\begin{equation} \text{NCA}_\phi : \mathbf{R}^{H\times W\times n} \rightarrow \mathbf{R}^{H\times W\times n}, \end{equation}
where $\phi = (\kappa_1, \kappa_2,\Tilde{W}_1, \Tilde{W}_2, \Tilde{b}_1, \Tilde{b}_2)$ denotes the set of trainable parameters.

\begin{figure}[t]
\centering
\includegraphics[width=\textwidth,page=1,trim=2.2cm 26cm 2.5cm 3cm,clip]{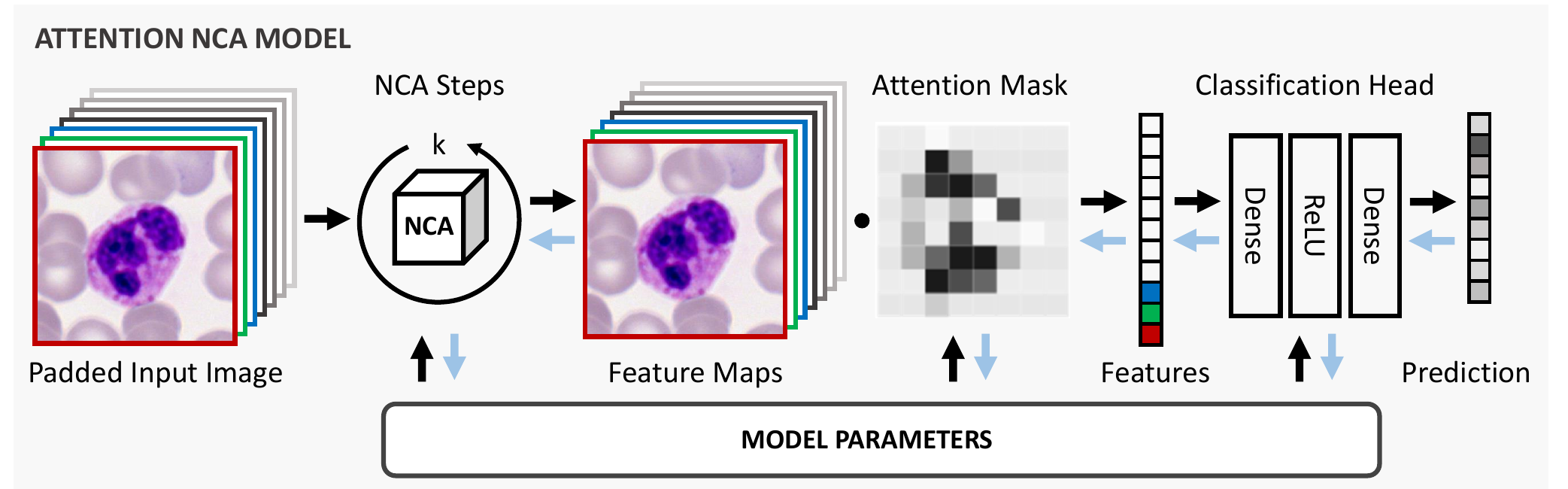}
\caption{Architecture of our attention neural cellular automata (aNCA). \textbf{A:} The model extracts feature maps from the padded input image using neural cellular automata. Each feature map is aggregated into a single value via a learnable weighted sum, a mechanism we call attention pooling, which enhances focus on the most informative regions. The resulting feature embedding is then passed through a fully connected network for classification. The gradient flow is indicated with blue arrows. The model including the attention mask and NCA is trained end-to-end.} 
\label{fig:fig1}
\end{figure}
\subsection{Classification Architecture}
After extracting feature channels using NCA, we aggregate each channel into a single value to form a feature embedding for classification. 
\begin{equation} \text{pool} : \mathbf{R}^{H\times W\times n} \rightarrow \mathbf{R}^n 
\end{equation}
To ensure a smooth transition between high and low dimensions, we implement attention pooling, which enables effective dimensionality reduction by learning an optimal weighting of the spatial features thus preserving crucial information.

\subsubsection{Attention Pooling}
We define a matrix of learnable weights $\theta \in \mathbf{R}^{H\times W}$, which is used to pool each of the feature channels $S^i$ into a weighted sum

\begin{equation} 
\text{pool} : \mathbf{R}^{H\times W\times n} \rightarrow \mathbf{R}^n,  \begin{pmatrix} S^1 \\ S^2 \\ \vdots \\ S^n \end{pmatrix} 
\mapsto \frac{1}{H\cdot W}
\begin{pmatrix} \Sigma_{i,j\in H,W}S^1_{i,j}\cdot \sigma(\theta_{i,j})  \\ \Sigma_{i,j\in H,W}S^2_{i,j}\cdot \sigma(\theta_{i,j}) \\ \vdots \\ \Sigma_{i,j\in H,W}S^n_{i,j}\cdot \sigma(\theta_{i,j}) \end{pmatrix},
\end{equation}
where \(\sigma(\theta_{i,j})=(1+e^{-\theta_{i,j}})^{-1}\).
In order to reduce noise in the pooling operation, we only consider a certain percentage of the highest activations in the weighted sum. We achieve this by sorting the products $S_{i,j}\cdot \sigma(\theta_{i,j})$, and averaging only the highest $10\%$ of values for each feature channel. This mechanism learns the most informative regions for classification.


\subsubsection{Classification Head}
To obtain the final class prediction, the aggregated feature embedding $v$ is passed through a two-layer fully connected network
\begin{equation}
    g:\mathbf{R}^n\rightarrow \mathbf{R}^C,\hspace{0.5cm}g(v) = \sigma (W_2\max(W_1v+b_1,0)+b_2),
\end{equation}
where $W_1$, $W_2$, $b_1$, $b_2$ are the weights and biases and $\sigma$ is a softmax activation. 

The learnable parameters of the model include the NCA parameters, the attention pooling weights, and the weights of the fully connected classification head. An overview of our method is illustrated in Figure \ref{fig:fig1}.
\section{Experiments \& Results}
\subsection{Datasets}
\begin{figure}[h]
\centering
\includegraphics[width=\textwidth,page=2,trim=0.4cm 22.5cm 0cm 0cm,clip]{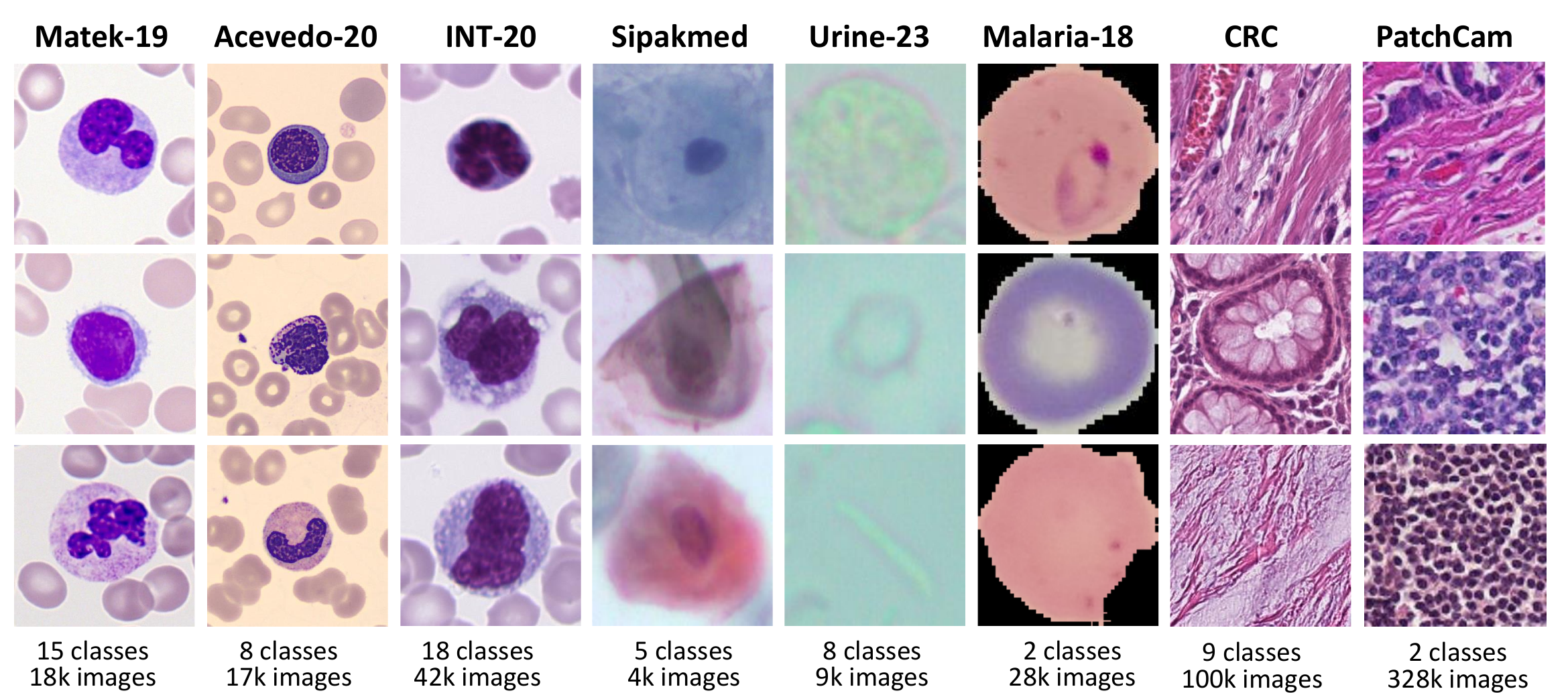}
\caption{Example images from the eight diverse microscopy datasets used in this study. The datasets cover a wide range of domains and differ in class distribution and size.} 
\label{fig:fig2}
\end{figure}

Eight diverse microscopy datasets are used for evaluation of our proposed method. Most of these datasets are publicly accessible. Example images from all datasets are displayed in Figure \ref{fig:fig2}.

\textbf{Matek-19}~\cite{matek2019human} consists of 18,365 expert-labeled white blood cell images sorted into 15 different classes. The samples were collected from 200 healthy donors and patients with acute myeloid leukemia at the Munich University Hospital. The images were captured at a resolution of $400\times 400$ pixels, corresponding to a physical size of $29\times 29$ micrometers.

\textbf{Acevedo-20}~\cite{acevedo2020dataset} contains 17,092 single-cell images sourced from healthy donors at the Hospital Clinic of Barcelona. The dataset consists of 8 cell classes, with each image having a resolution of $360\times 363$ pixels, corresponding to a scale of $36\times 36.3$ micrometers.

\textbf{INT-20} is an in-house dataset that contains approximately 42,000 single-cell images from 18 different classes. Each image has a resolution of $288 \times 288$ pixels, representing a field of view of $25 \times 25$ micrometers.

\textbf{Sipakmed}~\cite{plissiti2018sipakmed} consists of 4,049 manually annotated cell images extracted from Pap smear slides. The dataset contains 5 classes based on cytomorphology. The images were acquired using an optical microscope equipped with a CCD camera.

\textbf{Urine-23}~\cite{tuncer2023urine} comprises 8,509 labeled images containing 8 classes of urine sediment particles. The samples were collected from 409 patients at the Biochemistry Clinics of Elazig Fethi Sekin Central Hospital, captured with an Optika B293PLi microscope.

\textbf{Malaria-18}~\cite{rajaraman2018pre} contains 27,560 Giemsa stained red blood cell images sorted into parasitized and uninfected. The images were obtained from 193 patients at Chittagong Medical College Hospital, Bangladesh.

\textbf{CRC} ~\cite{kather2018100} is a collection of 100,000 image patches derived from histology slides of colorectal tissue, covering 9 classes of normal and cancerous samples. Each patch is $224\times 224$ pixels, with a spatial resolution of $0.5$ micrometers per pixel. To ensure consistency, images are color-normalized using Macenko’s method\cite{macenko2009method}. An additional set of 7,180 independent samples is used for validation.

\textbf{PatchCam} \cite{bejnordi2017diagnostic} contains 327,680 histopathology image patches extracted from lymph node tissue sections, divided into 262,144 training samples, and validation and test set both of 32,768 samples. Each sample has $96\times 96$ pixels and a binary label indicating the presence or absence of metastatic tissue.

\subsection{Implementation details}
In this section, we describe the training setups and implementation settings of aNCA and baselines.

\textbf{Input format} 3-channels RGB images, resized to $64 \times 64$ pixels and normalized with dataset-specific mean and standard deviation. For training, the images are augmented by random rotation and flipping.

\textbf{Model parameters} The model used for the experiments consists of 128 feature channels, iterated with 64 steps. The fully connected hidden layer and the classifier have a dimension of 128. The learnable attention map has the same size as the input image, which contains $64 \times 64$ pixels.

\textbf{Training} As most of the datasets are imbalanced in their classwise distribution, we use focal loss \cite{lin2020focalloss} for training. On the datasets Matek-19, Acevedo-20, INT-20, Sipakmed, Urine-23, and Malaria-18, the model is trained for 32 epochs with a batch size of 16. We use Adam optimizer with a learning rate of 0.0004, a $\beta_1$ of 0.9 and $\beta_2$ of 0.999 for training with an exponential learning rate decay with weight 0.9999. On the datasets CRC and PatchCam, the model is only trained for 10 and 5 epochs, respectively, because they converge faster with a bigger amount of data. We adapted the learning rate to 0.001 for CRC and 0.004 for PatchCam. 

\textbf{Metrics} We perform stratified five fold cross validation and report the accuracy score on the validation split on most of the datasets. On CRC and PatchCam, where the data splits are predefined, we computed the means and standard deviations from five independent runs, and we report balanced accuracy score on CRC to align with the published baseline. The accuracies on PatchCam are evaluated on the test dataset.

\textbf{Baselines} 
To compare our approach with existing lightweight architectures, we evaluated MobileNetV3 (mobilenetv3\_small\_050) and MobileViT (mobilevit\_xxs) using the TIMM library \cite{rw2019timm}. 
In preliminary experiments, we identified 0.003 as an effective learning rate for both models, which was subsequently used across all datasets. The optimizer, batch size, augmentation, and loss function were kept identical to the aNCA training settings to maintain consistency. The only exception was for CRC and PatchCam, where training was limited to 16 epochs due to faster convergence on these larger datasets.
We prioritized a consistent training pipeline to ensure a fair comparison, while acknowledging that additional hyperparameter tuning and model-specific optimizations could further improve baseline performance.
To support reproducibility, we provide the full codebase, including training and evaluation scripts, along with detailed instructions for reproducing our results at [anonymized link].

\subsection{Results} 
The model's performance is evaluated against WBC-NCA \cite{deutges2024neural}, MobileNetV3 \cite{howard2019searching} (a lightweight CNN), MobileViT\textsubscript{xxs} \cite{mehta2021mobilevit} (a lightweight ViT), and published state of the art for every dataset. Table \ref{tab:accuracy} shows the mean and standard deviation of accuracy of the mentioned methods, as well as the number of their parameters.  

\begin{table}[ht]
\caption{aNCA consistently outperforms lightweight baselines across all datasets. Classification accuracies of our proposed aNCA against WBC-NCA, MobileNetV3, MobileViT\textsubscript{xxs}, and dataset-specific baselines are reported across eight datasets, with their model sizes in parentheses.
Standard accuracy is reported for most datasets, while balanced accuracy is used for CRC to align with the published baseline. The highest accuracy among lightweight models (<1M parameters) for each dataset is highlighted in \textbf{bold}. 
}
\centering
\begin{tabular}{p{1.8cm}|P{1.7cm}|P{1.7cm}|P{1.7cm}|P{1.7cm}||P{1.9cm}}
\diagbox[width=1.9cm,height=1.2cm,innerleftsep=0.05cm,innerrightsep=0.05cm]{Dataset}{Method} & \textbf{aNCA} (89k) & WBC-NCA (86k) & CNN (580k) &ViT \quad (955k) & Published (21M-139M) \\ 
\hline
Matek-19 & \textbf{95.3$\pm$0.4} & 92.6$\pm$0.4 & 90.1$\pm$0.3 & 92.2$\pm$0.2 & 96.1~\cite{matek2019human} \\ 

Acevedo-20 & \textbf{92.1$\pm$0.3} &  89.8$\pm$0.7 & 79.6$\pm$0.6 & 85.6$\pm$0.3 & 96.2~\cite{acevedo2019recognition} \\ 

INT-20 & \textbf{92.9$\pm$0.2} & 88.0$\pm$0.3 & 86.3$\pm$0.4 & 89.7$\pm$0.3 & 88.7~\cite{salehi2022unsupervised} \\ 

Sipakmed & \textbf{94.3$\pm$1.1} & 92.5$\pm$0.7 & 82.9$\pm$2.0 & 87.8$\pm$1.7 &  99.1~\cite{KARRI2022797} \\ 

Urine-23 & \textbf{92.7$\pm$0.4} & 87.9$\pm$1.3 & 79.6$\pm$1.5 & 85.4$\pm$1.5 & 96.0~\cite{diagnostics13071299} \\

Malaria-18 & \textbf{96.7$\pm$0.2} &  96.5$\pm$0.3 & 93.4$\pm$1.7 & 95.2$\pm$0.4 & 98.9~\cite{rajaraman2018understanding} \\ 

CRC & \textbf{91.6$\pm$0.5} & 87.7$\pm$1.3 & 70.5$\pm$4.1 & 74.6$\pm$1.8 & 93.0~\cite{roth2024lowresource} \\

PatchCam & \textbf{83.5$\pm$0.3} & 82.5$\pm$0.2 & 73.6$\pm$0.9 & 73.9$\pm$0.8 & 85.4~\cite{roth2024lowresource} \\ 
\end{tabular}
\label{tab:accuracy}
\end{table}

\subsubsection{Published Baselines} 
Matek et al.~\cite{matek2019human} reported an accuracy of 96.1\% on Matek-19 with a model based on ResNeXt \cite{xie2017aggregated}. 
Salehi et al.~\cite{salehi2022unsupervised} trained the same model on INT-20, reporting an accuracy of 88.7\%. Acevedo et al.~\cite{acevedo2019recognition} report an accuracy of 96.2\% with a VGG-16 architecture \cite{simonyan2015deepconvolutionalnetworkslargescale} on Acevedo-20. Karri et al.~\cite{KARRI2022797} reached an accuracy of 99.12\% using a 19-layer CNN 
on Sipakmed~\cite{plissiti2018sipakmed}. Muhammed et al.~\cite{diagnostics13071299} proposed a hybrid model combining textural features with a ResNet50~\cite{He2015} 
backbone on Urine-23, achieving 96\% accuracy. Rajaraman et al.~\cite{rajaraman2018understanding} applied a VGG-16~\cite{simonyan2015deepconvolutionalnetworkslargescale} 
backbone on the Malaria-18 dataset with 98.9\% accuracy.
For the two pathology datasets, CRC and PatchCam, we use ViT-Small proposed by Roth et al.~\cite{roth2024lowresource} as the baseline. Specifically, we take the reported CRC results from the paper and reproduce the same setup to train on PatchCam.
To maintain consistency with their results, we present the balanced accuracy score for CRC, while standard accuracy is used for all other datasets.

\subsection{Ablation studies}
We conducted ablation studies based on two design choices. Firstly, we consider  
\textbf{different percentages of pixel values} used for averaging in the pooling operation. Our results indicate that selecting the top 10\% yields the highest accuracy across most datasets. Table \ref{tab:ablation} presents the model's performance using different percentages. 
Secondly, we evaluate an
\textbf{alternative implementation of the attention map}.
    In the proposed aNCA model, we define an array of learnable weights used in a weighted average. 
    Additionally, we explore an alternative approach where the attention map is generated by combining the extracted feature maps via a learnable convolutional layer. 
    The last column of Table \ref{tab:ablation} show its accuracies.

\begin{table}[ht]
\caption{Averaging the highest 10\% of pixels in feature maps achieves the highest performance on most datasets. For each NCA feature channel, we take the average of the highest $x$ percent of pixel values as the reduced feature. We studied the performance when setting $x$ to 5\%, 10\%, 20\%, and 50\% (Column 1-4). We also studied an alternative approach (conv) where the attention map is generated through a convolutional layer that operates on the feature (Column 5). The highest accuracy for each dataset is highlighted in \textbf{bold}, the second highest is \underline{underlined}.}
\centering
\begin{tabular}{p{1.8cm}|P{1.7cm}|P{1.7cm}|P{1.7cm}|P{1.7cm}|P{1.7cm}}
\diagbox[width=1.9cm,height=1.2cm,innerleftsep=0.05cm,innerrightsep=0.05cm]{Dataset}{Method} & aNCA \quad(5\%) &  \textbf{aNCA} (10\%) & aNCA (20\%) & aNCA (50\%) & aNCA \quad (conv,86k) \\ 
\hline
Matek-19 & \underline{95.2$\pm$0.3} & \textbf{95.3$\pm$0.4} & \textbf{95.3$\pm$0.3} & 95.1$\pm$0.3 & 94.8$\pm$0.3 \\ 

Acevedo-20 & \underline{91.8$\pm$0.2} & \textbf{92.1$\pm$0.3} & 91.6$\pm$0.3 & 91.3$\pm$0.3 & \underline{91.8$\pm$0.2}   \\ 

INT-20 & 92.8$\pm$0.2 & \underline{92.9$\pm$0.2} & \textbf{93.0$\pm$0.2} & 92.8$\pm$0.4 & 92.2$\pm$0.2  \\ 

Sipakmed & \underline{93.1$\pm$0.5} & \textbf{94.3$\pm$1.1} & 91.9$\pm$2.2 & 92.4$\pm$0.8  & 92.5$\pm$0.8\\ 

Urine-23 & \underline{92.0$\pm$0.5} & \textbf{92.7$\pm$0.4} & 90.9$\pm$0.2 & 89.4$\pm$1.2  & 90.4$\pm$0.4  \\

Malaria-18 & \underline{96.6$\pm$0.2} & \textbf{96.7$\pm$0.2} & 96.2$\pm$0.2 & 96.3$\pm$0.1 &  96.5$\pm$0.3   \\ 

CRC & 91.0$\pm$0.9 & \textbf{91.6$\pm$0.5} & \underline{91.4$\pm$0.3} & 90.7$\pm$0.6 & 91.1$\pm$0.6\\

PatchCam & \underline{83.0$\pm$0.4} & \textbf{83.5$\pm$0.3} & 82.4$\pm$0.6 & 82.2$\pm$0.5  & 81.3$\pm$1.0  \\ 
\end{tabular}
\label{tab:ablation}
\end{table}


\section{Conclusion} We introduced aNCA, a novel image classification model that enhances existing NCA models by incorporating attention pooling, resulting in significantly improved accuracy. By learning the optimal spatial weighting through an attention map, aNCA effectively transitions between high-dimensional NCA features and low-dimensional embeddings for the classification head. Additionally, aNCA effectively reduces noisy features by focusing only on the top 10\% values of each feature channel.

Evaluations on eight clinical microscopy datasets demonstrate that aNCA consistently achieves high accuracy across various modalities, outperforming both existing NCA models and other lightweight alternatives. Furthermore, aNCA requires substantially fewer parameters compared to conventional methods.

Our work highlights the potential of aNCA in microscopy image classification tasks. As an effective, robust, and explainable feature extractor, it is particularly well-suited for medical settings. 
While further work is needed to enhance computational efficiency and deployment readiness, our results suggest that NCA-based models may offer a promising alternative for flexible and interpretable approaches to medical image analysis.\\

\begin{credits}
\subsubsection{\ackname} C.M. acknowledges funding from the European Research Council (ERC)
under the European Union's Horizon 2020 research and innovation program (Grant Agreement No. 866411 \& 101113551 \& 101213822) and support from the Hightech Agenda Bayern.

\subsubsection{\discintname}
The authors have no competing interests to declare that are relevant to the content of this article.
\end{credits}

%
%
%
\bibliographystyle{splncs04}
\bibliography{Paper-106}

\begin{thebibliography}{10}
\providecommand{\url}[1]{\texttt{#1}}
\providecommand{\urlprefix}{URL }
\providecommand{\doi}[1]{https://doi.org/#1}

\bibitem{acevedo2019recognition}
Acevedo, A., Alf{\'e}rez, S., Merino, A., Puigv{\'\i}, L., Rodellar, J.: Recognition of peripheral blood cell images using convolutional neural networks. Computer methods and programs in biomedicine  \textbf{180},  105020 (2019)

\bibitem{acevedo2020dataset}
Acevedo, A., Merino, A., Alf{\'e}rez, S., Molina, {\'A}., Bold{\'u}, L., Rodellar, J.: A dataset of microscopic peripheral blood cell images for development of automatic recognition systems. Data in brief  \textbf{30} (2020)

\bibitem{bejnordi2017diagnostic}
Bejnordi, B.E., Veta, M., Van~Diest, P.J., Van~Ginneken, B., Karssemeijer, N., Litjens, G., Van Der~Laak, J.A., Hermsen, M., Manson, Q.F., Balkenhol, M., et~al.: Diagnostic assessment of deep learning algorithms for detection of lymph node metastases in women with breast cancer. Jama  \textbf{318}(22),  2199--2210 (2017)

\bibitem{deutges2024neural}
Deutges, M., Sadafi, A., Navab, N., Marr, C.: Neural cellular automata for lightweight, robust and explainable classification of white blood cell images. In: International Conference on Medical Image Computing and Computer-Assisted Intervention. pp. 693--702. Springer (2024)

\bibitem{He2015}
He, K., Zhang, X., Ren, S., Sun, J.: Deep residual learning for image recognition. arXiv preprint arXiv:1512.03385  (2015)

\bibitem{hehr2023explainable}
Hehr, M., Sadafi, A., Matek, C., Lienemann, P., Pohlkamp, C., Haferlach, T., Spiekermann, K., Marr, C.: Explainable ai identifies diagnostic cells of genetic aml subtypes. PLOS Digital Health  \textbf{2}(3),  e0000187 (2023)

\bibitem{howard2019searching}
Howard, A., Sandler, M., Chu, G., Chen, L.C., Chen, B., Tan, M., Wang, W., Zhu, Y., Pang, R., Vasudevan, V., et~al.: Searching for mobilenetv3. In: Proceedings of the IEEE/CVF international conference on computer vision. pp. 1314--1324 (2019)

\bibitem{KARRI2022797}
Karri, M., Annavarapu, C.S.R., Mallik, S., Zhao, Z., Acharya, U.R.: Multi-class nucleus detection and classification using deep convolutional neural network with enhanced high dimensional dissimilarity translation model on cervical cells. Biocybernetics and Biomedical Engineering  \textbf{42}(3),  797--814 (2022). \doi{10.1016/j.bbe.2022.06.003}

\bibitem{kather2018100}
Kather, J.N., Halama, N., Marx, A., et~al.: 100,000 histological images of human colorectal cancer and healthy tissue. Zenodo10  \textbf{5281}(9) (2018)

\bibitem{lin2020focalloss}
Lin, T.Y., Goyal, P., Girshick, R., He, K., Dollár, P.: Focal loss for dense object detection. IEEE Transactions on Pattern Analysis and Machine Intelligence  \textbf{42}(2),  318--327 (2020). \doi{10.1109/TPAMI.2018.2858826}

\bibitem{liu2025hasd}
Liu, J., Li, H., Yang, C., Deutges, M., Sadafi, A., You, X., Breininger, K., Navab, N., Sch{\"u}ffler, P.J.: Hasd: Hierarchical adaption for pathology slide-level domain-shift. In: International Conference on Medical Image Computing and Computer-Assisted Intervention. (2025)

\bibitem{macenko2009method}
Macenko, M., Niethammer, M., Marron, J.S., Borland, D., Woosley, J.T., Guan, X., Schmitt, C., Thomas, N.E.: A method for normalizing histology slides for quantitative analysis. In: 2009 IEEE international symposium on biomedical imaging: from nano to macro. pp. 1107--1110. IEEE (2009)

\bibitem{matek2019human}
Matek, C., Schwarz, S., Spiekermann, K., Marr, C.: Human-level recognition of blast cells in acute myeloid leukaemia with convolutional neural networks. Nature Machine Intelligence  \textbf{1}(11),  538--544 (2019)

\bibitem{mehta2021mobilevit}
Mehta, S., Rastegari, M.: Mobilevit: light-weight, general-purpose, and mobile-friendly vision transformer. arXiv preprint arXiv:2110.02178  (2021)

\bibitem{mordvintsev2020growing}
Mordvintsev, A., Randazzo, E., Niklasson, E., Levin, M.: Growing neural cellular automata. Distill  \textbf{5}(2), ~e23 (2020)

\bibitem{plissiti2018sipakmed}
Plissiti, M.E., Dimitrakopoulos, P., Sfikas, G., Nikou, C., Krikoni, O., Charchanti, A.: Sipakmed: A new dataset for feature and image based classification of normal and pathological cervical cells in pap smear images. In: 2018 25th IEEE international conference on image processing (ICIP). pp. 3144--3148. IEEE (2018)

\bibitem{rajaraman2018pre}
Rajaraman, S., Antani, S.K., Poostchi, M., Silamut, K., Hossain, M.A., Maude, R.J., Jaeger, S., Thoma, G.R.: Pre-trained convolutional neural networks as feature extractors toward improved malaria parasite detection in thin blood smear images. PeerJ  \textbf{6},  e4568 (2018)

\bibitem{rajaraman2018understanding}
Rajaraman, S., Silamut, K., Hossain, M.A., Ersoy, I., Maude, R.J., Jaeger, S., Thoma, G.R., Antani, S.K.: Understanding the learned behavior of customized convolutional neural networks toward malaria parasite detection in thin blood smear images. Journal of Medical Imaging  \textbf{5}(3),  034501--034501 (2018)

\bibitem{roth2024lowresource}
Roth, B., Koch, V., Wagner, S.J., Schnabel, J.A., Marr, C., Peng, T.: Low-resource finetuning of foundation models beats state-of-the-art in histopathology (2024)

\bibitem{sadafi2023redtell}
Sadafi, A., Bordukova, M., Makhro, A., Navab, N., Bogdanova, A., Marr, C.: Redtell: an ai tool for interpretable analysis of red blood cell morphology. Frontiers in Physiology  \textbf{14},  1058720 (2023)

\bibitem{sadafi2019multiclass}
Sadafi, A., Koehler, N., Makhro, A., Bogdanova, A., Navab, N., Marr, C., Peng, T.: Multiclass deep active learning for detecting red blood cell subtypes in brightfield microscopy. In: Medical Image Computing and Computer Assisted Intervention--MICCAI 2019: 22nd International Conference, Shenzhen, China, October 13--17, 2019, Proceedings, Part I 22. pp. 685--693. Springer (2019)

\bibitem{salehi2022unsupervised}
Salehi, R., Sadafi, A., Gruber, A., Lienemann, P., Navab, N., Albarqouni, S., Marr, C.: Unsupervised cross-domain feature extraction for single blood cell image classification. In: Medical Image Computing and Computer Assisted Intervention--MICCAI 2022: 25th International Conference, Singapore, September 18--22, 2022, Proceedings, Part III. pp. 739--748. Springer (2022)

\bibitem{simonyan2015deepconvolutionalnetworkslargescale}
Simonyan, K., Zisserman, A.: Very deep convolutional networks for large-scale image recognition (2015)

\bibitem{tesfaldet2022attention}
Tesfaldet, M., Nowrouzezahrai, D., Pal, C.: Attention-based neural cellular automata. Advances in Neural Information Processing Systems  \textbf{35},  8174--8186 (2022)

\bibitem{tuncer2023urine}
Tuncer, T., Erku{\c{s}}, M., {\c{C}}{\i}nar, A., Ayy{\i}ld{\i}z, H., Tuncer, S.A.: Urine dataset having eight particles classes. arXiv preprint arXiv:2302.09312  (2023)

\bibitem{rw2019timm}
Wightman, R.: Pytorch image models (2019), \url{https://github.com/rwightman/pytorch-image-models}

\bibitem{xie2017aggregated}
Xie, S., Girshick, R., Doll{\'a}r, P., Tu, Z., He, K.: Aggregated residual transformations for deep neural networks. In: Proceedings of the IEEE conference on computer vision and pattern recognition. pp. 1492--1500 (2017)

\bibitem{yang2025hierarchical}
Yang, C., Deutges, M., Navab, N., Sadafi, A., Marr, C.: Hierarchical neural cellular automata for lightweight microscopy image classification. In: International Conference on Information Processing in Medical Imaging. Springer (2025)

\bibitem{diagnostics13071299}
Yildirim, M., Bingol, H., Cengil, E., Aslan, S., Baykara, M.: Automatic classification of particles in the urine sediment test with the developed artificial intelligence-based hybrid model. Diagnostics  \textbf{13}(7) (2023). \doi{10.3390/diagnostics13071299}

\end{thebibliography}
\nocite{yang2025hierarchical}
%

\end{document}